\documentclass[mlabstract]{jmlr}

\usepackage{longtable}

\usepackage{booktabs}
\usepackage{siunitx} 
\usepackage{wrapfig}
\usepackage{caption}
\usepackage{placeins}

\usepackage{etoolbox}
\makeatletter
\preto\section{\par\parshape=0\relax}
\preto\subsection{\par\parshape=0\relax}
\preto\subsubsection{\par\parshape=0\relax}
\makeatother

\theorembodyfont{\upshape}
\theoremheaderfont{\scshape}
\theorempostheader{:}
\theoremsep{\newline}

\jmlrvolume{}
\firstpageno{1}

\jmlryear{2025}
\jmlrworkshop{Symmetry and Geometry in Neural Representations}

\title[Do traveling waves make good positional encodings?]{Do traveling waves make good positional encodings?}

\author{%
Chase van de Geijn$^{1,\dagger}$, Ayush Paliwal$^{1,2}$, Timo Lüddecke$^1$, Alexander S. Ecker$^{1,2}$\\
\addr{
$^1$Institute of Computer Science and Campus Institute Data Science, University of Göttingen\\
$^2$Max Planck Institute for Dynamics and Self-Organization, Göttingen, Germany\\
\footnotetext{$^\dagger$\texttt{chase.geijn@uni-goettingen.de}}
}}

\begin{document}

\maketitle

\begin{abstract}
Transformers rely on positional encoding to compensate for the inherent permutation invariance of self-attention. Traditional approaches use absolute sinusoidal embeddings or learned positional vectors, while more recent methods emphasize relative encodings to better capture translation equivariances. In this work, we propose RollPE, a novel positional encoding mechanism based on \emph{traveling waves}, implemented by applying a circular roll operation to the query and key tensors in self-attention. This operation induces a relative shift in phase across positions, allowing the model to compute attention as a function of positional differences rather than absolute indices. We show this simple method significantly outperforms traditional absolute positional embeddings and is comparable to RoPE. We derive a continuous case of RollPE which implicitly imposes a topographic structure on the query and key space. We further derive a mathematical equivalence of RollPE to a particular configuration of RoPE. Viewing RollPE through the lens of traveling waves may allow us to simplify RoPE and relate it to processes of information flow in the brain.
\end{abstract}

\section{Introduction}
The transformer architecture has achieved state-of-the-art performance across natural language processing, vision, and multimodal tasks. A central feature enabling this success is the self-attention mechanism, which models pairwise dependencies between tokens. However, because attention is inherently permutation-invariant, positional encoding is required to inject sequence order information.

Absolute encodings -- such as fixed sinusoidal embeddings \citep{vaswani2017attention} or learned vectors \citep{dosovitskiy2020image} -- assign each token a distinct position representation. While effective, these encodings lack \emph{relative awareness}: the model must learn to infer positional differences implicitly. This motivated the widely praised Rotary Positional Embeddings (RoPE) \citep{su2024roformer} which efficiently implements a relative positional encoding, or \textit{shift-equivariance}, by rotating the query and key vectors. However, recent evidence suggests strict equivariance is not a necessity for RoPE bringing into question what makes a good  positional encoding \citep{ostmeier2024liere, me, barbero2024round}.

We introduce \emph{rolling positional encodings} (RollPEs), a deliberately simplistic approach where position is encoded by \emph{rolling} query and key tensors before computing their dot product. This operation is rather trivially a relative positional encoding, but has a compelling interpretation as a traveling wave. These positional encodings induce a topographic arrangement in the query and key space as detailed by \citet{keller2021topographic}, and, by adding additional smoothness constraints over this topographic structure, lead to the spatial loss of TDANNs \citep{margalit2024unifying}, which have been shown to reproduce aspects of the hallmark behavior in the ventral stream in the visual systems of primates \citep{lee2020topographic}. Furthermore, within neuroscience there has been recent evidence that traveling waves play a significant role in the formation of long-term memory \citep{muller2018cortical} and that humans encode visual events as multiplexed traveling waves \citep{king2021human}. While we propose RollPE as a toy model, it links directly with these recent trends in theoretical neuroscience.

In our initial experiments, these simple embeddings significantly outperform classic encodings and,
through Multiplexed RollPE, we show once again -- as suggested by \citet{me} -- this behavior does not appear to be due to its relative inductive bias. These embeddings have a similar behavior to RoPE; in fact, we can mathematically derive a RollPE as a form of RoPE. We hypothesize that many of the properties and interpretations of RollPE apply transitively to RoPE. Thus, we hypothesize that RoPE's success may be derived from the implicit topographic structure and traveling waves that it and RollPE impose. Through this extended abstract, we motivate our on-going work into viewing the RoPE through the lens of traveling waves.

\section{Roll Positional Encoding}
Let a sequence of hidden states be represented as $X \in \mathbb{R}^{t \times n}$, with queries $Q = XW_Q$, keys $K = XW_K$, and values $V = XW_V$. In standard attention,
\begin{equation}
\text{Attn}(Q,K,V) = \text{softmax}\left(\frac{QK^\top}{\sqrt{d}}\right)V.
\end{equation}
\begin{minipage}{\linewidth}
We define the circular roll operator $\mathrm{Roll}_p:\mathbb{R}^n\to\mathbb{R}^n$ which maps \(
q_i \mapsto q_{i'},
\) where $i'=(i+p)\bmod n$. We could represent this in matrix form by defining the permutation matrix $S\in\mathbb{R}^{n\times n}$ be the 1-step roll matrix,

\begin{wraptable}[3]{r}{0.37\textwidth}
\vspace{-0.5cm}
\caption{Accuracy on CIFAR100}\label{tab:Results}
\centering
\footnotesize
\setlength{\tabcolsep}{4pt} %
\begin{tabular}{llcccccc}
\toprule
\textbf{PE}
& \textbf{CIFAR100} \\
\midrule
Baseline (APE)      &    64.2{\tiny$\pm0.9$}  \\
Axial RoPE    &  72.4{\tiny$\pm0.4$}   \\
\midrule
 RollPE       & 72.1{\tiny$\pm0.1$}     \\
 Multiplexed RollPE  & 73.4{\tiny$\pm0.5$}        \\
\bottomrule
\end{tabular}
\vspace{-2.0cm}
\end{wraptable}

\begin{equation}
     S^1 \;=\;
\begin{bmatrix}
0 & 1 & 0 & \cdots & 0 \\
0 & 0 & 1 & \cdots & 0 \\
\vdots & & \ddots & \ddots & \vdots \\
0 & \cdots & 0 & 0 & 1 \\
1 & 0 & \cdots & 0 & 0
\end{bmatrix}, \quad
\mathrm{Roll}_p\,(\mathbf{q}) \;=\;S^p\mathbf{q}.
\end{equation}
\vspace{.5pt}
\end{minipage}
As a positional encoding, we apply this to both queries and keys when calculating the attention score as done in \cite{su2024roformer},

\begin{equation}
\alpha_{i,j} = \frac{\mathrm{Roll}_{p_i}(\mathbf{q}_i) \cdot \mathrm{Roll}_{p_j}(\mathbf{k}_j)}{\sqrt{d}}.
\end{equation}

Note, the attention score between token $i$ and $j$ now depends on the relative displacement induced by the roll (proof in Appendix \ref{app:rollpe}). That is, RollPE is \textit{equivariant} to position. One can extend this simple rolling positional encoding by multiplexing -- i.\,e.\ representing queries as the superposition of multiple vectors which are rolled at different shift speeds. This gives Multiplexed RollPE (see Appendix \ref{app:multiplex} for details). While this yields better results, it requires more parameters and breaks equivariance. 

From the results in Table \ref{tab:Results}, we see that RollPE outperforms classic ViT positional encodings and performs similarly to RoPE. We also observe that Multiplexed RollPE outperforms RollPE, which can suggest that strict shift-equivariance may not be necessary as also suggested in \citet{me}.

\subsection{Beyond discrete positions}
One obvious flaw in RollPE is the need for discrete positions. While this is often reasonable in vision and language, it limits applications to continuous data such as point clouds. The second flaw is that RollPE is inherently periodic with period $n$. While this is not a problem for low-context domains such as vision---where images are typically represented with on the order of 16 patches in each direction \citep{dosovitskiy2020image} ---this becomes very limiting for language, where desirable context length is on the order of millions. Both problems can be addressed by generalizing the shift operator $S$ using its Lie algebra.

While $\mathrm{Roll}$ is only defined for integer shifts, we can write
\begin{equation} \label{eq:nabla}
\mathcal{A} := \log S,
\qquad
S = \exp(\mathcal{A}),
\end{equation}
where $\exp$ and $\log$ denote the matrix exponential and logarithm. Since $S$ is a permutation (and hence orthogonal) matrix, $\mathcal{A}$ belongs to the Lie algebra $\mathfrak{so}(n)$, i.\,e.\ it is skew-symmetric: $\mathcal{A}^\top = - \mathcal{A}$. This lets us define a continuous shift operator
\begin{equation} \label{eq:continroll}
\mathrm{Roll}_p(\mathbf{q}) \;=\; \exp\!\Bigl(\tfrac{p}{\lambda}\mathcal{A}\Bigr)\,\mathbf{q},
\end{equation}
which is now well-defined for all $p \in \mathbb{R}$ and allows the period to be stretched by a wavelength parameter $\lambda$. By changing the wavelength parameter, one can modulate the periodicity of RollPE.


\section{RollPE is RoPE}

Because $\mathcal{A}$ is guaranteed to be skew-symmetric, it can be decomposed into \begin{equation}
    \mathcal{A} = \mathbf{U} \mathbf{\Lambda}\mathbf{U}^\dagger,
\end{equation}
where $\mathbf{\Lambda}$ is purely imaginary, $\mathbf{U}$ is unitary, and $\dagger$ is the Hermitian transpose. Eq.~\ref{eq:continroll} can then be written \(\mathbf{U} \exp(\tfrac{p}{\lambda}\mathbf{\Lambda}) \mathbf{U}^\top\). In the case of the circular roll matrix, it is well known that $\mathbf{U}$ and $\mathbf{\Lambda}$ correspond to the discrete Fourier transform (DFT) matrices, $F$ and $\mathrm{diag}\!\bigl(2\pi i \frac{p k}{\lambda n}\bigr)_{k=0}^{n-1}$, respectively. Thus, the continuous Roll operator is given by
\begin{equation} \label{eq:fourier}
S(p) := F^\ast \,\mathrm{diag}\!\bigl(e^{2\pi i \frac{p k}{\lambda n}}\bigr)_{k=0}^{n-1}\, F,
\end{equation}
where $n$ is the dimension of the query vector and $k$ is the enumeration of its eigenvalues, and $F^\ast$ is the inverse Fourier matrix.
Notice, $\mathrm{diag}\!\bigl(e^{2\pi i \frac{p k}{\lambda n}}\bigr)_{k=0}^{n-1}$ is equivalent to a RoPE matrix.

\begin{theorem} RollPE can be represented as RoPE.
\end{theorem}
\proof{So long as Lie algebra given by Eq.~\ref{eq:nabla} is skew-symmetric RollPE is a special case of LieRE \citep{ostmeier2024liere}, which has been shown to be equivalent to a configuration of RoPE for one-dimensional positions \citep{me}. $\hfill \square$}

To be exact, RollPE corresponds to RoPE if the rotation frequency per dimension is $\omega_k = \frac{2\pi k}{\lambda n}$. The traditional formulation of RoPE uses fixed frequencies given by $\omega_k = 10000^{-2k/n} = e^{-2k/n \ln 10000}$. Note, one could define $\lambda = -\frac{\pi}{\ln 10000}$ to get these expressions close to each other -- however, the RoPE frequencies require an additional exponentiation. The behavior of this extra exponentiation is currently unknown, but to our understanding it is included by tradition rather than necessity.

\section{Discussion}

RollPE is an interesting case to study because other architectures have shown the Roll operation to induce equivariant capsules and reproduce topographic structuring similar to that observed in the visual systems of primates \citep{keller2021topographic, keller2021modeling}. One can abstract the query and key vectors of continuous RollPE as continuous signals over a continuous circle that are sampled at discrete ``sensors". By imposing structure on these signals one recovers topographics methods on a simplified one-dimensional cortical surface \citep{lee2019topographic, margalit2024unifying}.

\subsection{Topological Regularization}
In many settings, small variations in token position should not result in disproportionate changes in semantic representation. This means a token with a small shift to position should remain highly correlated with itself, i.\,e., if
\begin{equation}
\frac{\text{Roll}_p (\mathbf{q}) \cdot \text{Roll}_{p+\Delta p}(\mathbf{q})}{\|\mathbf{q}\|^2} \;\geq\; 1 - \epsilon ,
\end{equation}
then the representations are close in Euclidean space:
\begin{equation}
\|\mathbf{q} - \text{Roll}_{\Delta p}(\mathbf{q})\| < \epsilon.
\end{equation}

From a topological perspective, this constraint enforces that the signal defined over the underlying manifold varies smoothly with respect to positional shifts. In other words, we are encouraging the representation to be \emph{Lipschitz continuous} with respect to positional perturbations.

In practice, such smoothness can be promoted through an auxiliary regularization loss that penalizes differences between neighboring latent dimensions. This formulation directly parallels Laplacian regularization in graph-based learning \citep{kipf2016semi}, where our graph is simply a directed cyclic graph.

\paragraph{Topographic methods} In parallel, Topographic Deep Artificial NN (TDANN) models have sought to impose biologically inspired constraints to deep learning models by imposing a wiring length loss \citep{margalit2024unifying}. As a proxy for wiring length, they impose a spatial loss over the activations which are reshaped to a ``cortical sheet" -- which corresponds directly to the above regularization, but with a more complicated mesh than a cyclic graph. These models have been said to reproduce the neuronal clustering patterns associated with face recognition found in the visual cortex in primates \citep{lee2020topographic}. While their method can be seen as imposing a structure as a 2D square lattice -- as opposed to RollPEs cycle graph -- one can imagine generalizing traveling waves over more complex topographies. \citet{pang2023geometric} and \citet{horibe2019curved} have suggested that the topographic and geometric structure of the brain's connectome plays an important role in how waves propagate and the types of waveforms -- suggesting a benefit to extending the topological structuring of RollPE.

\paragraph{Conclusion} RoPE is strongly connected to traveling wave dynamics and we believe that it can be tied to many biological motivations. Because of its connection to RollPE, we hypothesize RoPE may perform well because it has similar properties to RollPE. That is, we hypothesize traveling wave dynamics is what makes RoPE a ``good" positional encoding.





\bibliography{NeurReps_2025_Template-3/biblio}

\appendix

\section{Related Work}

\paragraph{Neuro-Symbolic AI}
Using the Roll operation within deep learning to encode position has been proposed in \citet{keller2021topographic} to encode time for sequential data, such as rotating digits or color changes. They modeled a generation process that was motivated by predictive coding where the next frame of the sequence should be attained by rolling the latent vector. They found that these models, like the TDANNs \citep{lee2020topographic}, learn spatial organization and selectivity toward categories such as faces, similar to the Fusiform Face Area \citep{keller2021modeling}. This naturally led to more recent on Neural Wave Machines \citep{keller2023neural} and Kuramoto models \cite{miyato2024artificial}. 

The idea of Kuramoto models is to phase align with waves, so that ``neurons that wire together fire together". In many ways, the classic attention mechanism in transformers also encourages phase alignment between related tokens, meaning there is likely a direct relationship of RollPE to AKoRN \citep{miyato2024artificial}. Similarly, the line of papers \citep{lowe2022complex, lowe2023rotating} that led to AKoRN began with complex autoencoders, which are strikingly similar to RoPE. These ``synchrony models" \citep{reichert2013neuronal} try to answer the ``perceptual binding problem" of neuroscience of integrating features such as shape, color, and location into one object representation \citep{greff2020binding}. One answer for this problem is through hyperdimensional computing \citep{kanerva2009hyperdimensional} which has trended toward the direction of \textit{phasor} representations \citep{smolensky1990tensor, kymn2024computing} which, once again, are strikingly similar to the form or rotary encodings. Tracing these ideas back further, one arrives, once again, at traveling wave behavior in the brain \citep{eckhorn1988coherent}.

\section{RollPE is Relative} \label{app:rollpe}
One can write $\mathrm{Roll}_p(\mathbf{q})$ in matrix form as
\begin{equation}
    \mathbf{q}' =\mathrm{Roll}_p(\mathbf{q}) = S^p \mathbf{q},
\end{equation}
now let the attention score be given by \( \alpha(\mathbf{q}', \mathbf{k}') =\mathbf{q}'^\top\mathbf{k}'\) where $\mathbf{q}'$ and $\mathbf{k}'$ are the positionally encoded query and key vectors. The attention score can be expanded as,
\[ \alpha(\mathbf{q}', \mathbf{k}') =(S^{p_q}\mathbf{q})^\top S^{p_k}\mathbf{k}\]
where $p_q$ and $p_k$ are the positions of the query and key token. Because $S$ is a permutation -- and thus orthonormal -- we can write,
\[ \alpha(\mathbf{q}', \mathbf{k}') = \mathbf{q}^\top S^{p_k-p_q}\mathbf{k}.\]
Thus, the positional encoding only depends on the (signed) relative distance $p_k-p_q$.

\section{Multiplexed RollPE} \label{app:multiplex}

Multiplexing refers to the combination of multiple signals into a single composite signal—in this case, the superposition of several traveling waves. It is common in signal processing for communication devices, as well in neuroscience, e.\,g\,. in spiking neural network literature.

In \emph{Multiplexed RollPE}, the query/key representation is defined as a superposition of multiple components, each rolling at a different speed. 

Concretely, instead of a single query/key matrix, we introduce $\mathcal{W}$ distinct projections indexed by $w \in \{1, \dots, \mathcal{W}\}$:
\begin{equation}
    \mathbf{q}^{(w)} := W_Q^{(w)} X.
\end{equation}
The multiplexed positional encoding is then given by
\begin{equation}
    \mathrm{MPRoll}_p(\mathbf{q}) 
    := \sum_{w=1}^{\mathcal{W}} \mathrm{Roll}_{(wp)}\!\bigl(\mathbf{q}^{(w)}\bigr).
\end{equation}

\section{Experimental Setup}

We evaluate the positional encodings on CIFAR100 \citep{krizhevsky2009learning} using ViT-S \citep{dosovitskiy2020image}. The two positions for RollPE and Multiplexed RollPE are encoded axially where each coordinate affects a different sub-vector of the query/key analogously to what is done for Axial RoPE in \citep{heo2024rotary}.

\section{Exact Form of $\log S$}\label{app:logS}

The exact logarithm of the cyclic shift matrix $S \in \mathbb{R}^{n \times n}$ is most naturally expressed in the Fourier basis. Let $F$ be the discrete Fourier transform matrix. Then
\begin{equation} 
S^1 = F^\ast \,\mathrm{diag}\!\bigl(e^{2\pi i k/n}\bigr)_{k=0}^{n-1}\, F,
\end{equation}

so its logarithm is
\begin{equation}
\mathcal{A} = \log S \;=\; F^\ast \,\mathrm{diag}\!\Bigl(\tfrac{2\pi i k}{n}\Bigr)_{k=0}^{n-1}\, F.
\end{equation}

This operator is circulant and skew-Hermitian, not tridiagonal. Its action corresponds exactly to multiplication by $2\pi i k/n$ in the frequency domain, i.e.\ a discrete Fourier differentiation operator.  

The tridiagonal matrix shown in the main text should therefore be viewed as a \emph{local finite-difference approximation} to $\mathcal{A}$, which captures the intuition of a derivative with periodic boundary conditions, but is not the literal matrix logarithm. In the continuum limit, both agree with the true derivative, but at finite 
$n$, the Fourier-based logarithm preserves the entire spectrum exactly, while the tridiagonal stencil sacrifices spectral accuracy for sparsity and locality.

\section{RollPE is RoPE} \label{app:RoPERollPE}

In \citet{me}, they show that arbitrary dimensional special orthogonal transformations applied to queries and keys, $\mathbf{q}' = \exp({\mathcal{A}p}) \mathbf{q}$, can be decomposed into RoPE by taking the spectral decomposition of $\mathcal{A},$ where the rotation frequencies correspond to the eigenvalues. Since $\mathcal{A}$ can be seen as the generator $\mathcal{A},$ RollPE can be decomposed into RoPE. To be exact, RollPE is a particular case of RoPE where the rotation frequencies correspond to the frequencies of the discrete fourier differential operator given in Eq. \ref{eq:fourier}

\end{document}